\documentclass[letterpaper, 10 pt, conference]{ieeeconf} 
\IEEEoverridecommandlockouts
\usepackage{cite}
\usepackage{amsmath,amssymb,amsfonts}
\usepackage{algorithmic}
\usepackage{graphicx}
\usepackage{textcomp}
\usepackage{xcolor}
\usepackage{makecell}
\usepackage{romannum}
\usepackage{url}
\usepackage{dsfont}
\usepackage{listings}
\usepackage{fancyvrb}
\usepackage{framed}
\usepackage[listings,skins]{tcolorbox}
\usepackage[skipbelow=\topskip,skipabove=\topskip]{mdframed}
\mdfsetup{roundcorner=0}
\def\BibTeX{{\rm B\kern-.05em{\sc i\kern-.025em b}\kern-.08em
    T\kern-.1667em\lower.7ex\hbox{E}\kern-.125emX}}

    \newcommand*{\belowrulesepcolor}[1]{%
	\noalign{%
		\kern-\belowrulesep
		\begingroup
		\color{#1}%
		\hrule height\belowrulesep 
		\endgroup
	}%
}
\newcommand*{\aboverulesepcolor}[1]{%
	\noalign{%
		\begingroup
		\color{#1}%
		\hrule height\aboverulesep
		\endgroup
		\kern-\aboverulesep
	}%
}

\lstdefinestyle{customc}{
  belowcaptionskip=1\baselineskip,
  breaklines=true,
  frame=L,
  xleftmargin=\parindent,
  language=C,
  showstringspaces=false,
  basicstyle=\footnotesize\ttfamily,
  keywordstyle=\bfseries\color{green!40!black},
  commentstyle=\itshape\color{purple!40!black},
  identifierstyle=\color{blue},
  stringstyle=\color{orange},
}

\lstdefinestyle{customasm}{
  belowcaptionskip=1\baselineskip,
  frame=L,
  xleftmargin=\parindent,
  language=[x86masm]Assembler,
  basicstyle=\footnotesize\ttfamily,
  commentstyle=\itshape\color{purple!40!black},
}

\begin{document}

\title{fpgaDDS: An Intra-FPGA Data Distribution Service for ROS 2 Robotics Applications}

\author{Christian Lienen$^{1}$, Sorel Horst Middeke$^{1}$, and Marco Platzner$^{1}$%
\thanks{$^{1}$Christian Lienen, Sorel Horst Middeke, and Marco Platzner are with the Department of Computer Science,
Paderborn University, 33098 DE Paderborn, Germany
        {\tt\small christian.lienen@upb.de}, {\tt\small sorel.middeke@gmail.com}, {\tt\small platzner@upb.de}}%
}

\maketitle

\begin{abstract}
Modern computing platforms for robotics applications comprise a set of heterogeneous elements, e.g., multi-core CPUs, embedded GPUs, and FPGAs. FPGAs are reprogrammable hardware devices that allow for fast and energy-efficient computation of many relevant tasks in robotics. ROS is the de-facto programming standard for robotics and decomposes an application into a set of communicating nodes. ReconROS is a previous approach that can map complete ROS nodes into hardware for acceleration. Since ReconROS relies on standard ROS communication layers, exchanging data between hardware-mapped nodes can lead to a performance bottleneck.

This paper presents fpgaDDS, a lean data distribution service for hardware-mapped ROS 2 nodes. fpgaDDS relies on a customized and statically generated streaming-based communication architecture. We detail this communication architecture with its components and outline its benefits. We evaluate fpgaDDS on a test example and a larger autonomous vehicle case study. Compared to a ROS 2 application in software, we achieve speedups of up to 13.34 and reduce jitter by two orders of magnitude.
\end{abstract}

\section{Introduction}
\label{sec:Introduction}

The computation demand of robotics applications has increased in the last few years because modern robots gather more and more sensory information about their environment and employ sophisticated algorithms for decision-making. The rapid adoption of advanced machine learning for robotics additionally accelerates this trend. 

Modern robotics hardware architectures comprise a heterogeneous set of computing elements to provide the needed performance in an energy-efficient way. Computing elements include multi-core CPUs, embedded general-purpose GPUs, and FPGAs. FPGAs are reprogrammable hardware devices that are flexible since they can provide efficient acceleration for a wide range of workloads.

The Robot Operating System (ROS) is the de-facto standard for robotics applications. 
Following the design philosophy of ROS, a robotics application is decomposed into nodes, which implement sub-functions of the overall application. Such functional decomposition facilitates easier development, maintenance, and increased reusability of code. ROS nodes can leverage different types of communication for data exchange, 
e.g., many-to-many publish-subscribe communication where one or more nodes publish data on specific topics, and other nodes that subscribe to these topics receive the data.

Recently, the integration of FPGA-based hardware accelerators into ROS-based applications has been proposed. The presented approaches either integrate hardware accelerator kernels into ROS nodes and trigger kernel execution similar to remote procedure calls~\cite{Yamashina2016, reconfros} or move complete nodes into the hardware domain~\cite{lienen2021design}.

The latter approach is rather flexible and uses a standard ROS communication layer for data transfer. This, however, can induce significant overhead because all data passes shared main memory. Even for directly communicating ROS nodes in hardware, the data is transferred via main memory under the control of the software in the ROS communication layer. 

The novel contribution of this paper is the presentation of fpgaDDS, a lean data distribution service for ROS nodes that are mapped to hardware. fpgaDDS is based on a  customized streaming communication architecture and integrates into the ReconROS architecture and programming model~\cite{lienen2021design}. The presented approach leads to significant improvements in performance and reduction of jitter when executing ROS applications while at the same time maintaining the programming philosophy and functional decomposition model of ROS. 

The remainder of the paper is structured as follows: Section~\ref{sec:BackgroundRelatedWork} provides background, especially about ROS and hardware acceleration approaches for ROS. Section~\ref{sec:CommunicationMiddleware} elaborates on the novel intra-FPGA data distribution service fpgaDDS. Section~\ref{sec:Evaluation} reports on the experimental evaluation of fpgaDDS based on measurements. Section~\ref{sec:Conclusion} concludes the paper and provides an outlook on feature work.
\section{Background}
\label{sec:BackgroundRelatedWork}

In this section, we first introduce the Robot Operating System before we review hardware acceleration in ROS 2-based applications and ReconROS in more detail. 

\subsection{Robot Operating System}
\label{sec:BackgroundRelatedWork:ROS}

The Robot Operating System (ROS) comprises a set of software libraries, tools, and a methodology for the design of robotics applications. Following the design methodology of ROS, the overall application is decomposed into so-called ROS nodes. ROS nodes use different communication types to interact and exchange data. A topic-based publish-subscribe mechanism supports many-to-many communication in ROS. One-to-one communication follows a remote-procedure-call scheme leveraged by ROS services and actions. ROS applications are typically described by their computation graph, which comprises the ROS nodes and their communication as edges.

ROS 2 uses a message infrastructure for consistent information exchange between nodes. Messages are hierarchically specified using a simple definition language, starting from simple messages of built-in datatypes such as integers or floating point numbers up to complex multi-level message structures. Per default, the standard installation of ROS 2 provides a set of predefined messages and the option to define custom user-specified message types.

\begin{figure}[h]
    \centering
    \includegraphics[width=0.6\linewidth]{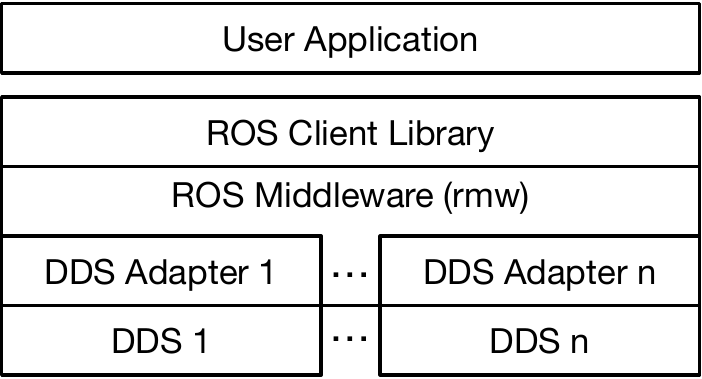}
    \caption{ROS 2 communication stack}
    \label{fig:background:ros_communication_stack}
\end{figure}

ROS 2 leverages standard communication middleware implementations following the Data Distribution Service (DDS) standard. Since DDS implementations differ slightly, the ROS middleware layer (rmw) provides individual adapters for each DDS implementation. In addition, the middleware layer provides communication interfaces for the ROS client library, which summarizes common client functionality~\cite{ros2}. Figure~\ref{fig:background:ros_communication_stack} shows the overall communication stack of ROS 2.

There are middlewares for intra-network communication and intra-platform communication. A prominent example of an intra-network communication middleware is FastDDS~\cite{FastRTPS} by eprosima. Iceoryx~\cite{Iceoryx} is an example of an intra-platform communication middleware that uses shared memory for data exchange. Further, there are also mixed forms that optimize communication depending on the destination. For instance, CycloneDDS~\cite{CycloneDDS} combines intra-network and intra-platform communication and employs Iceoryx internally.

\subsection{Hardware Acceleration for ROS applications}
\label{sec:BackgroundRelatedWork:HardwareAcceleration}

In recent years, hardware accelerators for ROS-based applications have been presented. Most works focus on ROS nodes, identify runtime-extensive parts and move them to a hardware accelerator. At runtime, the remaining ROS node calls the hardware accelerator like a remote-procedure call and possibly blocks during accelerator execution. Examples of this approach are described in~\cite{Yamashina2016,reconfros}, or are leveraged by industry products, e.g., Xilinx KRIA~\cite{mayoral2021kria}. 

There is also research to accelerate communication. In~\cite{Sugata2017}, the authors reduce communication overhead by separating control and data communication traffic handled in hardware. Follow-up work~\cite{8823798} proposed to interpret ROS messages with high-level hardware synthesis tool-flows, which reduces development time and creates more compact code.

In~\cite{mayoral2022robotcore}, the authors present two approaches for reducing communication 
overheads between nodes: kernel fusion and dedicated streaming queues. Kernel fusion merges acceleration kernels of different ROS nodes into one combined acceleration kernel. Dedicated streaming queues connect acceleration kernels using data-transmission interfaces. Both approaches break with ROS' node decomposition methodology, i.e., the original ROS computational graph gets transformed. 

FPGA-ROS~\cite{Podlubne2020} allows for implementing ROS nodes and their communication interfaces altogether in hardware. A streaming network connects application-specific hardware cores to a shared central Ethernet gateway, which realizes TCP/IP communication in hardware. A central manager instance supervises the overall application.

\subsection{ReconROS}
\label{sec:BackgroundRelatedWork:ReconROS}

ReconROS~\cite{lienen2021design} combines the reconfigurable hardware operating system ReconOS~\cite{Luebbers_Platzner_2009} with ROS 2 and provides a framework for hardware acceleration of ROS 2 applications. 

\begin{figure}[h]
    \centering
    \includegraphics[width=0.75\linewidth]{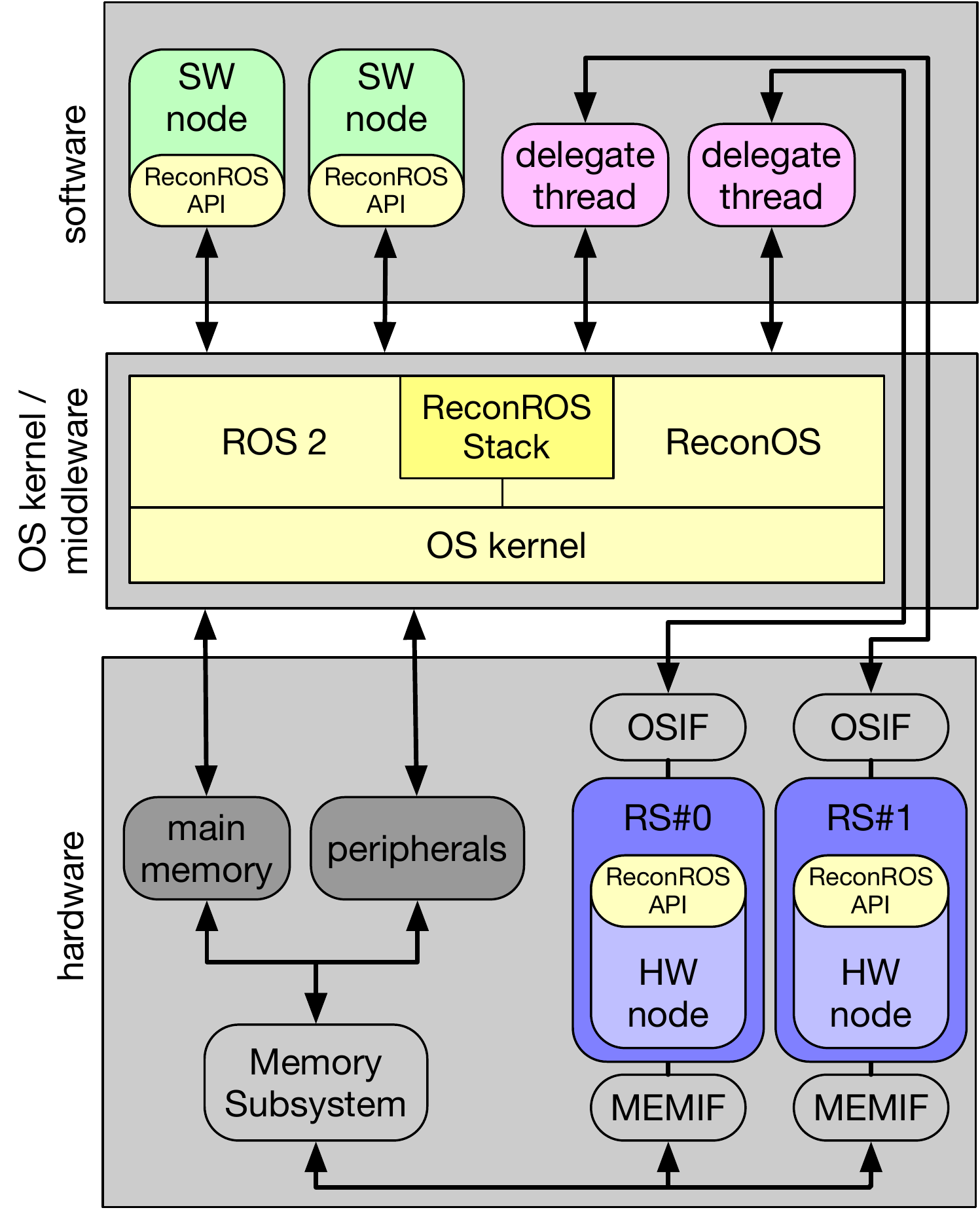}
    \caption{ReconROS Architecture (from~\cite{reconros_taskmapping}) }
    \label{fig:background:reconros_architecture}
\end{figure}

Figure~\ref{fig:background:reconros_architecture} shows the ReconROS architecture implemented on a platform FPGA comprising a multi-core CPU and a reconfigurable fabric. ROS 2 nodes implemented in hardware (HW nodes) are mapped to reconfigurable slots (RS), which are rectangular areas on the reconfigurable fabric. HW nodes use a memory interface (MEMIF) to access the shared virtual memory space, including memory-mapped peripherals. Further, each HW node relies on an operating system interface (OSIF) to connect to its delegate thread, a lightweight software thread that performs operating system calls on behalf of the HW node.

ReconROS extends a guest operating system, e.g., Linux, by ReconOS and ROS 2 and allows for the execution of ROS 2 nodes in hardware and software through the ReconROS stack and the ReconROS API. The ReconROS stack extends ReconOS by ROS-related objects, e.g., ROS publishers or subscribers. The ReconROS API is available for both SW and HW node implementations. It ensures a consistent programming model that even allows for dynamically changing the mapping of a ROS node to software or hardware or assigning a HW node to a specific reconfigurable slot at runtime. In this paper, however, we only discuss static mappings. 

For instance, a HW node that wants to receive a new message from a ROS topic sends the corresponding command via its OSIF to its delegate thread. The delegate receives the command,  relays it to the ReconROS stack, and waits in a blocked state until lower ROS layers provide a new message. This wakes up the delegate that then sends the address of the new message via the OSIF to the HW node. Finally, the HW node can read the message using its MEMIF. ReconROS builds on a shared memory architecture where all HW nodes share one access port to the main memory. Hence, the HW nodes share the bandwidth for the memory port, and together with the multi-core CPU, they share the total bandwidth to the main memory.

The ReconROS build flow starts from a configuration file that captures the information about the target platform, synthesis toolchains, ROS version, number of reconfigurable slots, the HW nodes with their ROS-specific resources, e.g., subscribers and publishers, and the mapping of HW nodes to reconfigurable slots and generates all required software binaries and the FPGA configuration bitstream. ReconROS supports the design of HW nodes using either the hardware description languages VHDL and Verilog or high-level synthesis from C/C++. 
\section{Intra-FPGA Data Distribution Service}
\label{sec:CommunicationMiddleware}

The ReconROS framework uses ROS 2 standard communication middlewares for data transport between nodes, including HW nodes. Although this excels in flexibility, as HW nodes can communicate with arbitrary SW and HW nodes, the overall application performance may be suboptimal due to the HW nodes' mechanism of calling ROS functions via their delegate threads and due to competing memory accesses of all HW nodes. 

This paper presents fpgaDDS, a novel and lean intra-FPGA data distribution service for ReconROS applications. fpgaDDS maps ROS communication between HW nodes to the reconfigurable fabric and thus avoids many memory accesses while maintaining the ROS programming model for standard ROS 2 publish-subscribe communication, i.e., no changes to the ROS nodes are required. Figure~\ref{fig:background:ros_communication_stack} sketches how fpgaDDS and the corresponding 
ReconROS DDS adapter integrates into the ROS 2 communication stack.

The new intra-FPGA DDS has two benefits: First, mapping the data transport between communicating HW nodes to the reconfigurable fabric saves many memory accesses and thus improves application performance. Second, moving nodes and communication from software to parallel executing hardware further reduces application execution time and jitter, which helps achieve predictable real-time behavior under ROS 2.

\begin{figure}[h]
    \centering
    \includegraphics[width=0.85\linewidth]{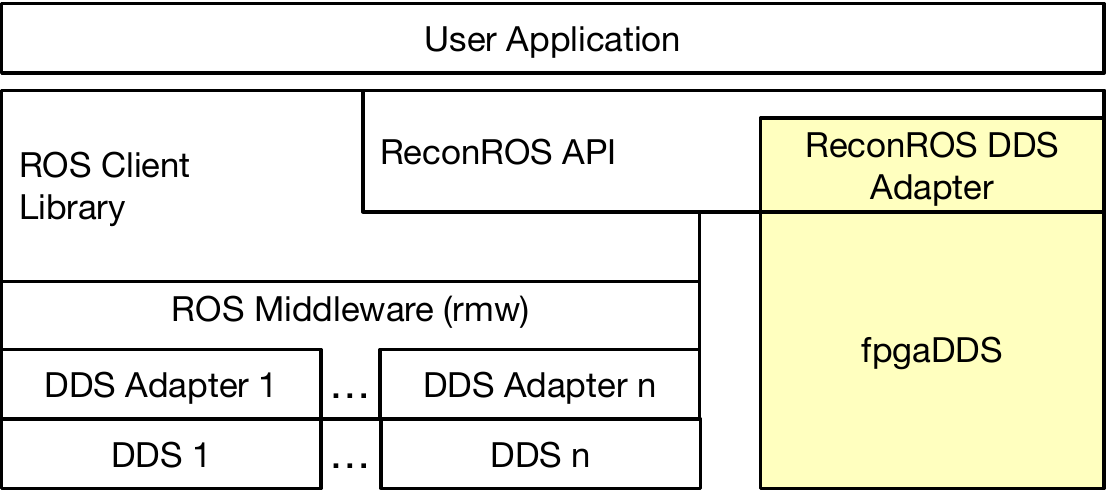}
    \caption{Extensions of the ReconROS communication stack include fpgaDDS and the corresponding ReconROS DDS adapter}
    \label{fig:CommunicationMiddleware:stack}
\end{figure}

\subsection{Intra-FPGA Communication Architecture}
\label{sec:CommunicationMiddleware:Infrastructure}

The publish-subscribe communication principle of ROS 2 inspires the structural design of the fpgaDDS communication architecture. The fpgaDDS-extended ReconROS build flow generates an application-specific static AXI-streaming (AXIS) network for each ROS topic separately. Such AXIS-based hardware-mapped topics (HMT) are lean, resulting in relatively simple communication protocols executed during runtime and high-performance implementations due to the total available bandwidth per HMT.

\begin{figure}[h]
    \centering
    \includegraphics[width=0.8\linewidth]{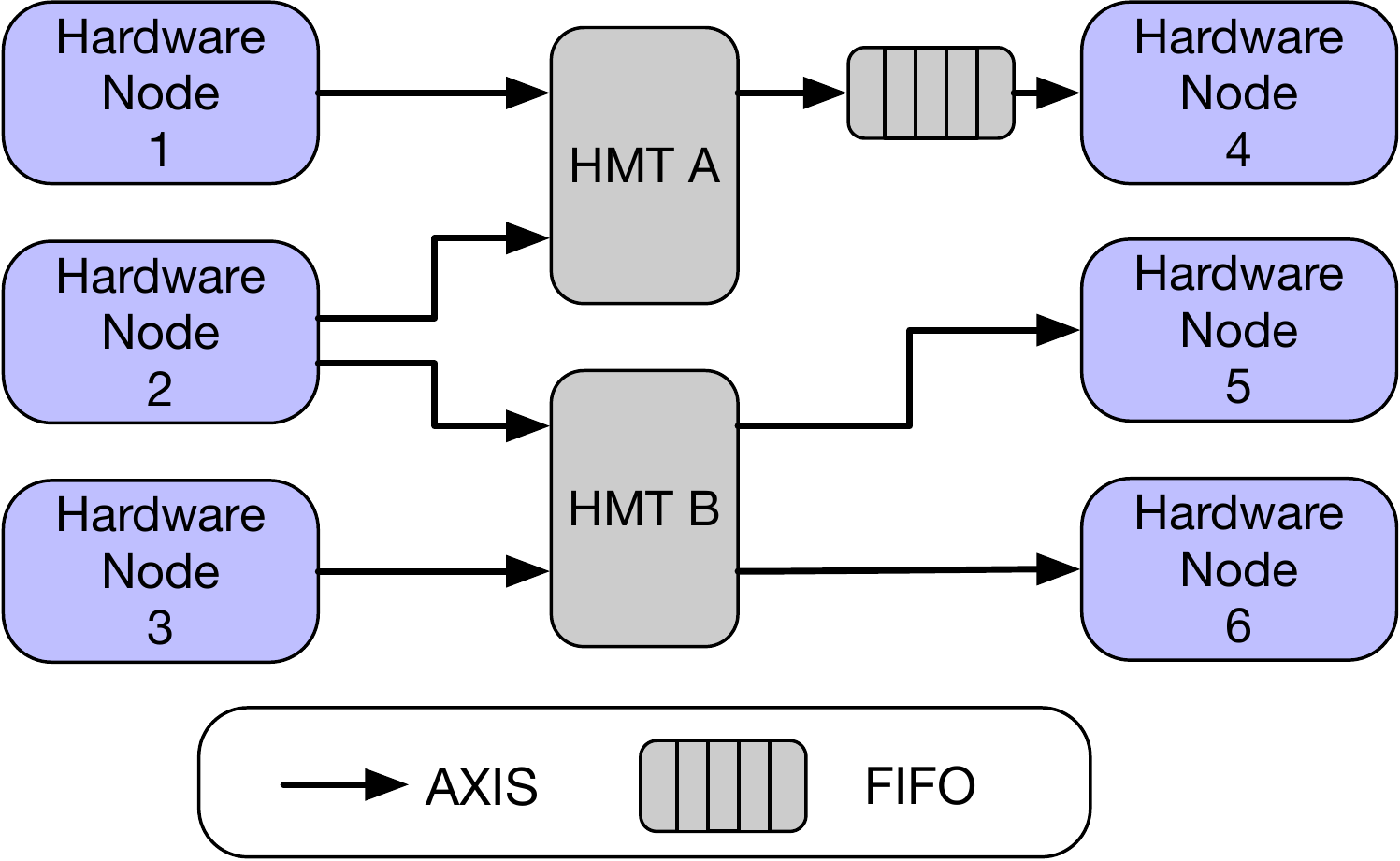}
    \caption{Example for an instance of the communication architecture with two hardware-mapped topics A and B}
    \label{fig:CommunicationMiddleware:example_architecture}
\end{figure}

Figure~\ref{fig:CommunicationMiddleware:example_architecture} shows a fpgaDDS example 
comprising six ReconROS HW nodes connected to two HMTs, A and B, by AXIS. In the figure, a directed connection from a HW node to a HMT represents a publisher, and a connection from a HMT to a HW node is a subscription. 

The internal design of a HMT depends on the required number of publishers and subscribers. In the simplest case, where the HMT receives data from one publisher and provides that data to one subscriber, the HMT includes a simple AXIS connection from input to output. In case there are more publishers for an HMT, the inputs are gathered by an AXI Interconnect IP block which arbitrates the input messages. The AXI-interconnect is configured to implement an arbitration based on complete messages, ensuring consistency in the sense that always complete messages are forwarded. If there are more HMT subscribers, the message is broadcast to all subscribers by an AXI Broadcast IP block. Naturally, AXI Interconnect and Broadcast IP blocks can be concatenated to realize HMTs with arbitrary numbers of publishers and subscribers.

Figure~\ref{fig:CommunicationMiddleware:example_architecture} also indicates a FIFO buffer for the subscription of topic A by HW node 4. Such FIFO buffers are optional components for subscribers of HMTs. By adding FIFO buffers, the data processing of the subscriber nodes can be decoupled from the topic. This behavior can be helpful for asynchronous communication, for example, where the subscriber reads the data from the topic at a different time. 

In ROS 2, DDS layers allow developers to specify various Quality-of-Service (QoS) parameters. fpgaDDS, with its AXIS streaming architecture, realizes, by default, publish-subscribe communication with the QoS parameters {\it Keep All} and {\it Reliable}, where no messages are discarded. The number of stored messages depends on the size of the FIFO. If the FIFO runs full, transmission blocks. In addition, due to its static DDS structure, fpgaDDS provides infinite {\it Lifespan Duration} and infinite {\it Lease Duration}.

The communication architecture for the fpgaDDS is synthesized based on the ReconROS project configuration file during design time. First, all subscribers and publishers with a hardware property attribute are identified and grouped for their topics. Then, for each HMT, the ReconROS build flow extends the HW nodes by input (subscription) and output (publishing) ports, inserts the HMT AXIS infrastructure, and connects the ports and the infrastructure accordingly. The overall procedure corresponds to the discovery procedure in standard DDS implementations. 

\subsection{ReconROS DDS Adapter}
\label{sec:CommunicationMiddleware:ProgrammingModel}

The ReconROS DDS adapter aims to close the gap between the programming model for streaming networks at the fpgaDDS layer and the publish/subscribe communication mechanism in ReconROS. As a result, the programmer can use similar blocking and non-blocking functions for interaction with software-mapped and hardware-mapped topics. 

In order to maintain compatibility between standard message structures provided by ROS 2 and the message structures used in our streaming fpgaDDS communication architecture, which is generated by a high-level synthesis tool flow, we need to serialize message objects on the transmitter side and de-serialize them on the receiver side.

Due to the multi-layer architecture of ROS 2 and the support of several different communication middlewares, the ROS 2 framework already provides functionality for ROS message serialization in software, even for complex nested message types. However, these functions can not be directly used for hardware generation since Vivado HLS provides limited support for pointer arithmetic, casting, and recursion. 

Therefore, we extract the structure of messages, i.e., their components and data types, from the message definition file and replace all non-primitives with their sub-components. This also allows for resolving multi-level nested messages. The resulting list of primitives and arrays is then transformed into a C macro for message publishing and subscribing. During hardware synthesis, these macros are in-lined and generate hardware to write and read data sequentially to and from AXI streaming interfaces in blocking and non-blocking versions.

\subsection{Execution Modes}
\label{sec:CommunicationMiddleware:ExecutionMode}

\begin{figure}[h]
    \centering
    \includegraphics[width=1.0\linewidth]{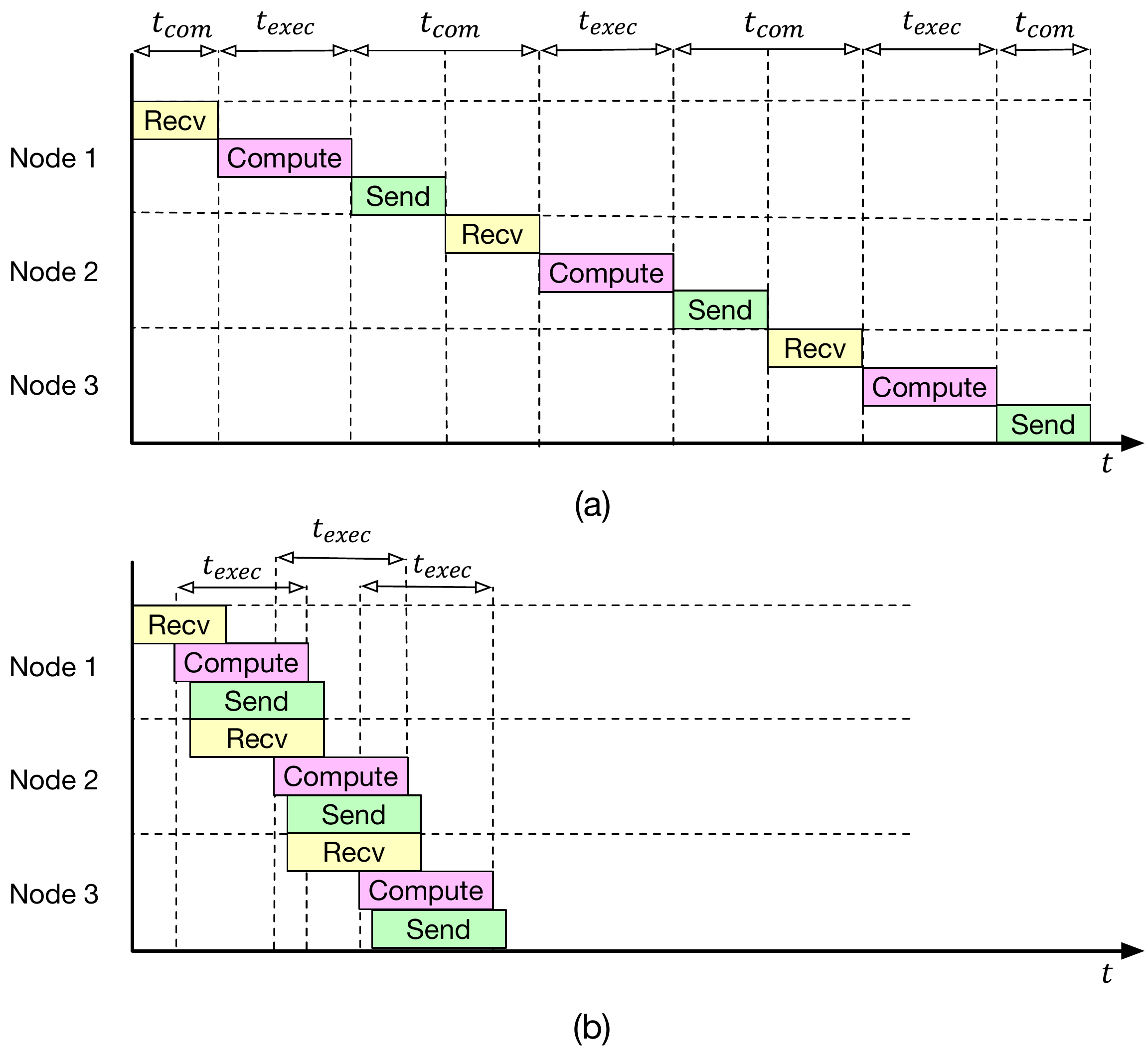}
    \caption{Execution modes for HW nodes with communication ($t_{com})$ and executio ($t_{exec})$ phases: (a) sequential execution and (b) dataflow execution}
    \label{fig:CommunicationMiddleware:execution_mode}
\end{figure}

Communicating HW nodes can be operated in two execution modes with fpgaDDS. The first mode is the same as when using a software-based DDS and operates the receive, compute, and send phases sequentially. Figure~\ref{fig:CommunicationMiddleware:execution_mode}-a shows an example of a chain of three nodes with publish-subscribe communication.

With fpgaDDS, we can additionally leverage the dataflow option of the high-level synthesis tool to create an implementation where the phases are overlapped as much as possible, constrained only by the data flow. This operation mode is exemplified in Figure~\ref{fig:CommunicationMiddleware:execution_mode}-b and can substantially improve overall execution time. This mode is useful for sets of nodes that operate on data in a streaming manner, which is typical, for example, for many low-level image processing tasks.

\section{Evaluation}
\label{sec:Evaluation}

This section reports on experiments with fpgaDDS. First, we describe the experimental setup, followed by experiments with HW nodes to demonstrate the benefit of using fpgaDDS. Finally, we present measurements on a real-world application of autonomous driving. 

\subsection{Experimental Setup}
\label{sec:Evaluation:Setup}

All HW nodes have been implemented in C++ and synthesized to hardware with the high-level synthesis tool Xilinx Vitis 2021.2. The computing platform is a ZCU104 evaluation board comprising an UltraScale+ MPSoC FPGA running Ubuntu Linux 18.04, ReconROS, and ROS 2 dashing. The ReconROS infrastructure and the HW nodes run at 100 MHz. For the autonomous driving example, we additionally use a desktop PC with an Intel Core i5-8000 CPU running Ubuntu Linux 18.04 and ROS 2 dashing, connected to the FPGA evaluation board via Gigabit Ethernet.

\subsection{Hardware-Mapped Topic Evaluation}
\label{sec:Evaluation:Synthetic}

In the first experiment, we implemented two HW nodes connected by a topic and measured the time it took from the start of sending to the end of receiving. Data transfers have been repeated 1000 times, and the average values are reported. Table~\ref{table:Experiments:communication_times} presents the resulting transfer times $t_{avg}$ and their standard deviation $\sigma$ for HW nodes using the software-mapped CycloneDDS and the hardware-mapped fpgaDDS for different message sizes. The table also lists the speedups achieved by fpgaDDS. The speedups show that the overhead for using CycloneDDS dominates for small data sizes. However, the speedup converges close to $2$ for larger data sizes because data has to be transferred twice over the MEMIF compared to one needed transmission via the HMT. However, the measurements for CycloneDDS represent the most optimistic case. With higher utilization of the MEMIF, the transmission time increases as the nodes share MEMIFs bandwidth.

Increasing the number of publishers and subscribers increases the transfer time for CycloneDDS. For fpgaDDS, the transfer time only increases with more publishers since the HMT arbitrates the incoming messages. However, adding more subscribers in fpgaDDS does not influence the transfer time.

\begin{table}[ht]
  \begin{center}
      \begin{tabular}{|r r r r|}
      \hline
      \makecell[r]{Message Size}       & \makecell[c]{CycloneDDS\\$t_{avg}$ $(\sigma)$ [ms]}  & \makecell[c]{fpgaDDS\\$t_{avg}$ $(\sigma)$ [ms]} & \makecell[c]{Speedup}\\[0.5ex] 
      \hline
      \hline
      \makecell[r]{3 kB}     & 0.06 (0.11)      & $<$0.01 ($<$0.01)  & 17.21 \\
      \makecell[r]{12 kB}    & 0.07 (0.06)      & 0.02 ($<$0.01)     & 4.71 \\
      \makecell[r]{50 kB}    & 0.19 (0.08)      & 0.06 ($<$0.01)     & 3.11 \\
      \makecell[r]{196 kB}   & 0.63 (0.04)      & 0.24 ($<$0.01)     & 2.54 \\
      \makecell[r]{786 kB}   & 2.34 (0.03)      & 0.98 ($<$0.01)     & 2.38 \\
      \makecell[r]{3146 kB}  & 9.24 (0.04)      & 3.93 ($<$0.01)     & 2.35 \\     
      \hline
      \end{tabular}
   \end{center}
  \caption{Communication times and speedup for CycloneDDS and fpgaDDS for different message sizes}
  \label{table:Experiments:communication_times}
\end{table}

\subsection{Autonomous Vehicle Example}
\label{sec:Evaluation:AutonomousVehicleExample}

\begin{figure*}[t]
	\centering
    \includegraphics[width=1.0\textwidth]{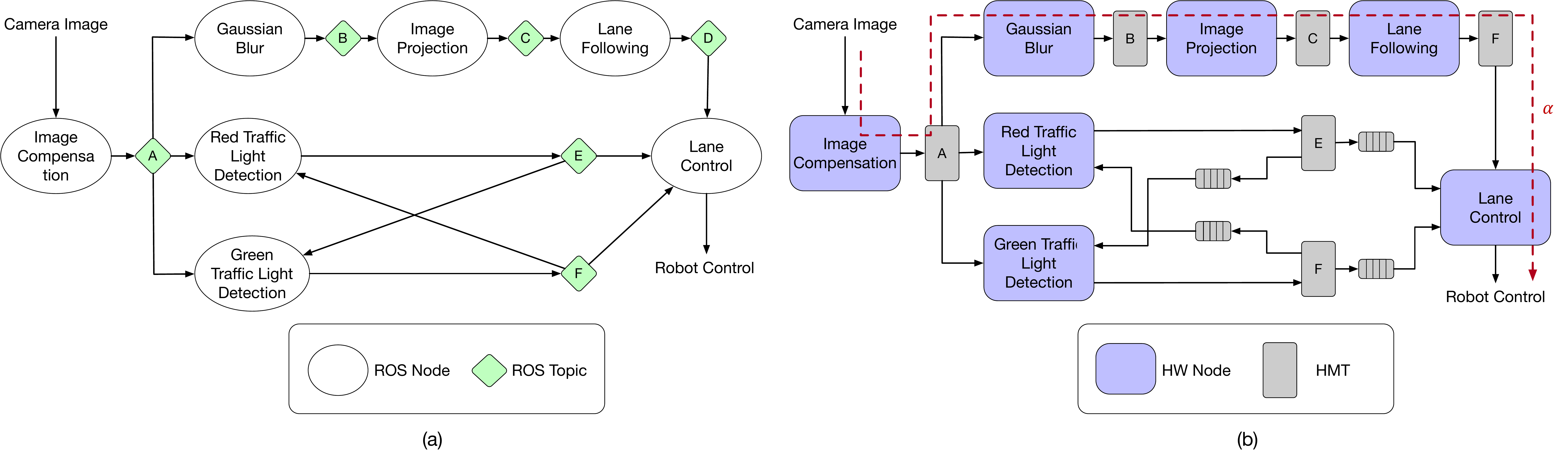}    
    \caption{Computation graph (a) and the resulting architecture graph (b) for the autonomous vehicle example}
    \label{fig:example_robotics_architecture}
\end{figure*}

In this experiment, we elaborate and extend the autonomous driving application from~\cite{reconros_taskmapping}. This application is inspired by the Turtlebot auto race challenge, follows a street lane, and handles traffic lights during the drive. 
The target robot is the Turtlebot 3 platform comprising a 0.3-megapixel camera for lane and traffic light detection. In our setup, the resulting ReconROS application runs on the FPGA evaluation board, connected to the desktop PC executing a Gazebo physics simulation environment. 

Figure~\ref{fig:example_robotics_architecture}-a displays the computational graph for the application. The only input of the computation graph is the camera image of the robot. The {\em Image Compensation} node calculates a histogram of the input image and uses it to remove outlier pixels. The resulting image is then published to {\em Topic A}. The {\em Gaussian Blur} node subscribes to the image data, applies a low-pass filter, and forwards the result to the image projection node through {\em Topic B}. The {\em Image Projection} node applies a warp transformation for an orthogonal view of the lane, then resizes the image to 1000x600 pixels. Finally, the {\em Lane Following} node extracts the lane's left and right stripes from image data and calculates the center point between them. This center point is sent to the node {\em Lane Control} that calculates driving commands from the center point.

In addition to the {\em Gaussian Blur} node, the {\em Red Traffic Light Detection} and the {\em Green Traffic Light} nodes subscribe to {\em Topic A}. Both nodes transform the image into the HSV colorspace and then detect a red or green traffic light. If a red traffic light is detected, the node publishes a stop command to {\em Topic E}, which leads to a stop of the robot by the lane control node. Additionally, that command activates green traffic light detection. When a green light is detected, the node publishes a start command to {\em Topic F}, which activates the lane control node again.

We have implemented the ROS 2 application of Figure~\ref{fig:example_robotics_architecture}-a in three different versions: (i) Software, where nodes run on the CPU based on CycloneDDS for communication, (ii) hardware, where all nodes are implemented using high-level synthesis but still use CycloneDDS for communication, and (iii) hardware, where the nodes use our novel fpgaDDS. The architecture graph for the third version is sketched in Figure~\ref{fig:example_robotics_architecture}-b. In addition, the subscribers of HMT {\em Topic E} and {\em F} are equipped with buffers to enable asynchronous communication via these HMTs.

The software implementation does not use any reconfigurable logic resources. Table~\ref{table:Experiments:Ressources_Example_Architecture} summarizes the resource utilization of the reconfigurable fabric for the hardware implementations. The table lists the resource types lookup-tables (CLB LUTs), dedicated memory blocks (BRAM, URAM), and arithmetic function blocks (DSP). The main insight from Table~\ref{table:Experiments:Ressources_Example_Architecture} is that fpgaDDS does not require more logic resources than the version with CycloneDDS. This is because the AXIS-based communication architecture requires relatively few logic resources. CycloneDDS, on the other hand, demands substantially more CLB LUTs for implementing in-node buffers. 

\begin{table}[ht]
  \begin{center}
      \begin{tabular}{|l r r r r|}
      \hline
      \makecell[c]{Implementation}       & CLB LUTs  & \makecell[c]{BRAM} & \makecell[c]{URAM} & \makecell[c]{DSPs}\\[0.5ex] 
      \hline
      \hline
      \makecell[c]{Hardware\\(CycloneDDS)}   & \makecell[c]{134252\\(58\%)} &  \makecell[c]{250.5\\(80\%)} & \makecell[c]{95\\(99\%)}  &  \makecell[c]{230\\(13\%)}\\
      \hline
      \makecell[c]{Hardware\\(fpgaDDS)}      &  \makecell[c]{44939\\(20\%)} &  \makecell[c]{255.0\\(82\%)} & \makecell[c]{91\\(95\%)}  &  \makecell[c]{230\\(13\%)}\\
      \hline
      \end{tabular}
   \end{center}
  \caption{Resource utilization of the hardware implementations (\% of the used XCZU7EV-2FFVC1156)}
  \label{table:Experiments:Ressources_Example_Architecture}
\end{table}

To evaluate the performance of the different communication middlewares, we have measured the execution times of the node chain $\alpha$ indicated as a red dashed line in Figure~\ref{fig:example_robotics_architecture}. The node chain starts with the image compensation node and ends with the lane control node. We have selected this chain of nodes since it subsumes most nodes regularly executed in the lane-following mode of the vehicle and feeds the topics periodically with message data. In contrast, 
topics {\em E} and {\em F} receive data only in the event of detected red or green traffic lights.

We have determined the execution of time node chain $\alpha$ by calculating the difference between the publishing time of the lane control node and the subscription time of the image compensation node. Again, we repeat the experiments 1000 times and report on the average and the standard deviation in Table~\ref{table:Experiments:Speedup_Example_Architecture}. Additionally, the table presents the speedups achieved over a pure software implementation. For example, implementing the ROS nodes in hardware but keeping the software-based CycloneDDS results in a speedup of 2.48. Mapping also communication to hardware with fpgaDDS gives a speedup of 13.34. Furthermore, with fpgaDDS, the standard deviation, which relates to jitter, reduces by two orders of magnitude.

\color{black}
\begin{table}[ht]
  \begin{center}
      \begin{tabular}{|l r r|}
      \hline
      \makecell[c]{Implementation}   &  $t_{avg}$ ($\sigma$) [ms] & Speedup  \\[0.5ex] 
      \hline
      \hline
      \makecell[c]{Software\\(CycloneDDS)}   & 274.44 (14.96)   &  1.00 \\
      \hline
      \makecell[c]{Hardware\\(CycloneDDS)}   & 110.72 (07.73)    &  2.48 \\ 
      \hline
      \makecell[c]{Hardware\\(fpgaDDS)}      & 20.58 (00.13)   & 13.34  \\ %
      \hline
      \end{tabular}
   \end{center}
  \caption{Execution times and speedups for node chain $\alpha$ of Figure~\ref{fig:example_robotics_architecture} and different implementation versions}
  \label{table:Experiments:Speedup_Example_Architecture}
\end{table}

\section{Conclusion and Future Work}
\label{sec:Conclusion}
This paper has presented fpgaDDS, a lean data distribution service for ROS nodes mapped to hardware. fpgaDDS is a statically generated, customized streaming communication architecture integrated into ReconROS that ensures low latency and high bandwidth communication between hardware nodes. Experiments have demonstrated significantly improved performance and reduced jitter. 

Future work is planned along two lines: First, we aim to extend fpgaDDS for ROS 2 services and actions and support a richer set of Quality-of-Service parameters.
Second, we want to make available fpgaDDS for dynamically scheduled ROS 2 nodes. ROS 2 foresees a so-called executor that dispatches callback functions on specific events for dynamic scheduling. On FPGA platforms, dynamic scheduling is supported by partial dynamic reconfiguration. Combining such dynamic techniques with fpgaDDS poses challenges since fpgaDDS is currently statically generated.

\bibliographystyle{IEEEtran}
\bibliography{lienen23_iros}

\begin{thebibliography}{10}
\providecommand{\url}[1]{#1}
\csname url@rmstyle\endcsname
\providecommand{\newblock}{\relax}
\providecommand{\bibinfo}[2]{#2}
\providecommand\BIBentrySTDinterwordspacing{\spaceskip=0pt\relax}
\providecommand\BIBentryALTinterwordstretchfactor{4}
\providecommand\BIBentryALTinterwordspacing{\spaceskip=\fontdimen2\font plus
\BIBentryALTinterwordstretchfactor\fontdimen3\font minus
  \fontdimen4\font\relax}
\providecommand\BIBforeignlanguage[2]{{%
\expandafter\ifx\csname l@#1\endcsname\relax
\typeout{** WARNING: IEEEtran.bst: No hyphenation pattern has been}%
\typeout{** loaded for the language `#1'. Using the pattern for}%
\typeout{** the default language instead.}%
\else
\language=\csname l@#1\endcsname
\fi
#2}}

\bibitem{Yamashina2016}
K.~Yamashina, H.~Kimura, T.~Ohkawa, K.~Ootsu, and T.~Yokota, ``{CReComp:
  Automated Design Tool for {ROS}-Compliant {FPGA} Component},'' in \emph{Proc.
  IEEE 10th International Symposium on Embedded Multicore/Many-Core
  Systems-on-Chip, MCSoC 2016}.\hskip 1em plus 0.5em minus 0.4em\relax IEEE,
  2016, pp. 138--145.

\bibitem{reconfros}
M.~Eisoldt, S.~Hinderink, M.~Tassemeier, M.~Flottmann, J.~Vana, T.~Wiemann,
  J.~Gaal, M.~Rothmann, and M.~Porrmann, ``{ReconfROS: Running ROS on
  Reconfigurable SoCs},'' in \emph{Proc.~2021 Drone Systems Engineering and
  Rapid Simulation and Performance Evaluation: Methods and Tools
  Proceedings}.\hskip 1em plus 0.5em minus 0.4em\relax ACM, 2021, p. 16–21.

\bibitem{lienen2021design}
C.~Lienen and M.~Platzner, ``{Design of Distributed Reconfigurable Robotics
  Systems with ReconROS},'' \emph{ACM Trans. Reconfigurable Technol. Syst.},
  vol.~15, no.~3, dec 2022.

\bibitem{ros2}
S.~Macenski, T.~Foote, B.~Gerkey, C.~Lalancette, and W.~Woodall, ``Robot
  operating system 2: Design, architecture, and uses in the wild,''
  \emph{Science Robotics}, vol.~7, no.~66, 2022.

\bibitem{FastRTPS}
``eprosima fast dds,'' \url{https://github.com/eProsima/Fast-DDS}, accessed:
  2023-02-28.

\bibitem{Iceoryx}
``iceoryx - true zero-copy inter-process-communication,''
  \url{https://github.com/eclipse-iceoryx/iceoryx}, accessed: 2023-02-28.

\bibitem{CycloneDDS}
``Eclipse cyclone dds,''
  \url{https://github.com/eclipse-cyclonedds/cyclonedds}, accessed: 2023-02-28.

\bibitem{mayoral2021kria}
V.~Mayoral-Vilches, ``{Kria Robotics Stack},''
  \url{https://www.xilinx.com/applications/industrial/robotics/wp540-kria-robotics-stack.html},
  2021, accessed: 2022-01-13.

\bibitem{Sugata2017}
Y.~Sugata, T.~Ohkawa, K.~Ootsu, and T.~Yokota, ``{Acceleration of
  Publish/Subscribe Messaging in {ROS}-Compliant {FPGA} Component},'' in
  \emph{Proc. of the 8th International Symposium on Highly Efficient
  Accelerators and Reconfigurable Technologies (HEART2017)}.\hskip 1em plus
  0.5em minus 0.4em\relax ACM, 2017.

\bibitem{8823798}
T.~{Ohkawa}, Y.~{Sugata}, H.~{Watanabe}, N.~{Ogura}, K.~{Ootsu}, and
  T.~{Yokota}, ``{High Level Synthesis of {ROS} Protocol Interpretation and
  Communication Circuit for {FPGA}},'' in \emph{Proc. 2019 IEEE/ACM 2nd
  International Workshop on Robotics Software Engineering (RoSE)}, 2019, pp.
  33--36.

\bibitem{mayoral2022robotcore}
V.~Mayoral-Vilches, S.~M. Neuman, B.~Plancher, and V.~J. Reddi, ``Robotcore: An
  open architecture for hardware acceleration in ros 2,'' in \emph{2022
  IEEE/RSJ International Conference on Intelligent Robots and Systems (IROS)},
  2022, pp. 9692--9699.

\bibitem{Podlubne2020}
A.~{Podlubne} and D.~{G\"ohringer}, ``{FPGA-ROS: Methodology to Augment the
  Robot Operating System with {FPGA} Designs},'' in \emph{Proc.~International
  Conference on ReConFigurable Computing and {FPGA}s (ReConFig)}, 2019.

\bibitem{Luebbers_Platzner_2009}
E.~L\"ubbers and M.~Platzner, ``{ReconOS: Multithreaded Programming for
  Reconfigurable Computers},'' \emph{ACM Transactions on Embedded Computing
  Systems}, vol.~9, no.~1, pp. 8:1--8:33, 2009.

\bibitem{reconros_taskmapping}
C.~Lienen and M.~Platzner, ``Task mapping for hardware-accelerated robotics
  applications using reconros,'' in \emph{2022 Sixth IEEE International
  Conference on Robotic Computing (IRC)}, 2022, pp. 148--155.

\end{thebibliography}

\end{document}